\documentclass{article}

\usepackage{microtype}
\usepackage{graphicx}
\usepackage{subfigure}
\usepackage{booktabs} %
\usepackage[usenames, dvipsnames]{color}
\usepackage{tabularx}
\usepackage{multirow}
\usepackage{amsfonts}
\usepackage{amsmath}
\usepackage{bbm}
\usepackage{enumitem}

\usepackage{hyperref}

\definecolor{orange}{RGB}{255, 125, 125}

\newcommand{\alphazero}{\emph{AlphaZero}}
\newcommand{\muzero}{\emph{MuZero}}
\newcommand{\smuzero}{\emph{Sampled MuZero}}
\newcommand{\ipi}{\mathcal{I}\pi}
\newcommand{\sample}{\beta}
\newcommand{\ibpi}{\hat{\mathcal{I}}_{\sample}\pi}
\newcommand{\ihpi}{\mathcal{I}\hat{\pi}_\sample}
\newcommand{\dmcs}{DeepMind Control Suite }
\newcommand{\rwrl}{Real-World RL }

\usepackage[accepted]{icml2020}

\icmltitlerunning{Learning and Planning in Complex Action Spaces}

\begin{document}

\twocolumn[
\icmltitle{Learning and Planning in Complex Action Spaces}

\icmlsetsymbol{equal}{*}

\begin{icmlauthorlist}
\icmlauthor{Thomas Hubert}{equal,dm}
\icmlauthor{Julian Schrittwieser}{equal,dm}
\icmlauthor{Ioannis Antonoglou}{dm}
\icmlauthor{Mohammadamin Barekatain}{dm}
\icmlauthor{Simon Schmitt}{dm}
\icmlauthor{David Silver}{dm}
\end{icmlauthorlist}

\icmlaffiliation{dm}{DeepMind, London, UK}

\icmlcorrespondingauthor{Thomas Hubert}{tkhubert@google.com}

\icmlkeywords{Machine Learning, ICML, RL, MCTS, Planning}

\vskip 0.3in
]

\printAffiliationsAndNotice{\icmlEqualContribution} %

\begin{abstract}
Many important real-world problems have action spaces that are high-dimensional, continuous or both, making full enumeration of all possible actions infeasible. Instead, only small subsets of actions can be sampled for the purpose of policy evaluation and improvement. In this paper, we propose a general framework to reason in a principled way about policy evaluation and improvement over such sampled action subsets. This sample-based policy iteration framework can in principle be applied to any reinforcement learning algorithm based upon policy iteration. Concretely, we propose \smuzero{}, an extension of the \muzero{} algorithm that is able to learn in domains with arbitrarily complex action spaces by planning over sampled actions. We demonstrate this approach on the classical board game of Go and on two continuous control benchmark domains: \dmcs and \rwrl Suite.

\end{abstract}

\section{Introduction}

Real-world environments abound with complexity in their action space. Physical reality is continuous both in space and time; hence many important problems, most notably physical control tasks, have continuous multi-dimensional action spaces. The joints of a robotic hand can assume arbitrary angles; the acceleration of a self-driving car should vary smoothly to minimise discomfort for passengers. Discrete problems also often have high-dimensional action spaces, leading to an exponential number of possible actions. Many other domains have richly structured actions spaces such as sentences, queries, images, or serialised objects. Consequently, a truly general reinforcement learning (RL) algorithm must be able to deal with such complex action spaces in order to be successfully applied to those real-world problems.

Recent advances in deep learning and RL have indeed led to remarkable progress in model-free RL algorithms for continuous action spaces \cite{lillicrap2015continuous,schulman2017proximal,d4pg,mpo,hoffman2020acme} and other complex action spaces \cite{dulacarnold2016deep}. Simultaneously, planning based methods have enjoyed huge successes in domains with discrete action spaces, surpassing human performance in the classical games of chess and Go \cite{Silver18AZ} or poker \cite{brown2018superhuman,deepstack}. The prospect of combining these two areas of research holds great promise for real-world applications.

The model-based \muzero{} \cite{muzero} RL algorithm took a step towards applicability in real-world problems by learning a model of the environment and thus unlocking the use of the powerful methods of planning in domains where the dynamics of the environment are unknown or impossible to simulate efficiently. However, \muzero{} was only applied to domains with relatively small action spaces; small enough to be in fact enumerated in full by the tree-based search at its core.

Sample-based methods provide a powerful approach to dealing with large complex actions spaces. Rather than enumerating all possible actions, the idea is to sample a small subset of actions and compute the optimal policy or value function with respect to those samples. This simple strategy is so general that it can be applied to large, continuous, or structured action spaces. Specifically, action sampling can be used both to propose improvements to the policy at each of the sampled actions, and subsequently to evaluate the proposed improvements. However, to correctly improve or evaluate the policy across the entire action space, and not just the samples, one must understand how the sampling procedure interacts with both policy improvement and policy evaluation.

In this work, we propose a framework to reason in a principled way about policy improvement and evaluation computed over small subsets of sampled actions. We show how this local information can be used to train a global policy, act and even perform explicit steps of policy evaluation for the purpose of planning and local policy iteration. This sample-based framework can in principle be applied to any reinforcement learning algorithm based upon policy iteration. Concretely, we propose \smuzero{}, an algorithmically simple extension of the \muzero{}\footnote{The discussion in this paper applies equally to \alphazero{} and \muzero{}; in the text we will only refer to \muzero{} for simplicity.} algorithm that facilitates its application to domains with complex action spaces.

To demonstrate the generality of this approach, we apply our algorithm to two continuous control benchmark domains, the \dmcs \cite{tassa2018deepmind} and \rwrl Suite \cite{dulacarnold2020empirical}. We also demonstrate that our algorithm can be applied to large discrete action spaces, by sampling the actions in the game of Go, and show that high performance can be maintained even when sub-sampling a small fraction of possible moves.

\section{Related Work}

Previous research in reinforcement learning for complex or continuous action spaces has often focused on model-free algorithms.

Deep Deterministic Policy Gradient (DDPG) exploits the fact that in action spaces that are entirely continuous (no discrete action dimensions), the action-value function $Q(s, a)$ can be assumed to be differentiable with respect to the action $a$ in order to efficiently compute policy gradients \cite{ddpg2014silver,lillicrap2015continuous}. Distributed Distributional Deterministic Policy Gradients (D4PG) extends DDPG by using a distributional value function and a distributed training setup \cite{d4pg}.
Trust Region Policy Optimisation (TRPO) uses a hard KL constraint to ensure that the updated policy remains close to the previous policy during the policy improvement step \cite{schulman2015trust}, to avoid catastrophic collapse.
Proximal Policy Optimisation (PPO) has the same goal as TRPO, but instead uses the KL-divergence as a penalty in the loss function or clipping in the value function \cite{schulman2017proximal}. This results in a simpler algorithm with empirically better performance.
In the regime of data-efficient off-policy algorithms, recent advances have derived actor-critic algorithms that optimise a (relative-)entropy regularised RL objective such as SAC \cite{haarnoja2018soft}, MPO \cite{mpo}, AWR \cite{peng2019advantage}. Among these, MPO uses a sample based policy improvement step that can be related to our algorithm (see section \ref{expectation-ipi}). Distributional MPO (DMPO) extends MPO to use a distributional Q-function \cite{hoffman2020acme}.

Model-based control for high dimensional action spaces has recently seen a resurgence of interest (see e.g. \cite{byravan2020imagined,hafner:planet,hafner2019dream, koul2020dream}). While most of these algorithms consider direct policy optimisation against a learned model some have considered combinations of rollout based search/planning with policy learning. \cite{piche2018probabilistic} use planning via sequential importance sampling of action sequences sampled from a SAC policy. \cite{bhardwaj2020information} use a learned simulator to construct K-step returns for learning a soft Q-function. Closest to our work, \cite{springenberg2020local} consider a sample based policy update similar to ours - but using a policy improvement operator based on the KL regularised objective rather than the MCTS based policy improvement that we consider here.

Sparse sampling algorithms \cite{kearns-mansour-ng:sparse-sampling} are an effective approach to planning in large state spaces. The main idea is to sample $K$ possible state transitions from each state, drawn from a generative model of the underlying MDP. Collectively, these samples provide a search tree over a subset of the MDP; planning over the sampled tree provides a near-optimal approximation, for large $K$, to the optimal policy for the full MDP, independent of the size of the state space. Indeed, sampling is known to address the curse of dimensionality in some cases \cite{rust1997}. However, sparse sampling typically enumerates all possible actions from each state, and does not address issues relating to large action spaces. In contrast, our method samples actions rather than state transitions. In principle, it would be straightforward to combine both ideas; however, we focus in this paper upon the novel aspect relating to large action spaces and utilise deterministic transition models.

There have been several previous attempts at generalising \alphazero{} and \muzero{} to continuous action spaces. These attempts have shown that such an extension is possible in principle, but have so far been restricted to very low dimensional cases and not yet demonstrated effectiveness in high-dimensional tasks.
A0C \cite{moerland2018a0c} describes an extension of \alphazero{} to continuous action spaces using a continuous policy representation and REINFORCE \cite{williams1992simple} to estimate the gradients for the reverse KL divergence between the neural network policy estimate and the target MCTS policy, demonstrating some learning on the 1D Pendulum task.
\cite{yang2020continuous} describe a similar extension of \muzero{} to continuous actions and show promising results outperforming soft actor-critic (SAC) \cite{haarnoja2018soft} on environments with 1 and 2 dimensional action spaces.

The factorised policy representation described by \cite{tang2020discretizing} shows good results in a variety of domains; by representing each action dimension with a separate categorical distribution it efficiently avoids the exponential explosion in the number of actions faced by a simple discretisation scheme.

\section{Background}
We consider a standard reinforcement learning setup in which an agent acts in an environment by sequentially choosing actions over a sequence of time-steps in order to maximise a cumulative reward. We model the problem as a Markov decision process (MDP) which comprises a state space $\mathcal{S}$, an action space $\mathcal{A}$, an initial state distribution, a stationary transition dynamics distribution and a reward function $r: \mathcal{S}\times\mathcal{A}\to\mathbb{R}$.

The agent's behaviour is controlled by a policy $\pi: \mathcal{S}\to\mathcal{P}(\mathcal{A})$ which maps states to a probability distribution over the action space. The return from a state is defined as the sum of discounted future rewards $G_t = \sum_i \gamma^{i} r(s_{t+i}, a_{t+i})$ where $\gamma$ is a discount factor in $[0,1]$. The goal of the agent is to learn a policy which maximises the expected return from the start distribution.

In order to do so, a common strategy called \emph{policy evaluation} consists in learning a value function that estimates the expected return of following policy $\pi$ from a state $s_t$ or a state action pair $(s_t, a_t)$. The value function can then be used in a process called \emph{policy improvement}, to find and learn better policies by for instance increasing the probabilities of actions with higher values. The process of repeatedly doing policy evaluation followed by policy improvement is at the heart of many reinforcement learning algorithms and is called \emph{policy iteration}.

Naturally, a lot of research focuses on improving the methods for \emph{policy evaluation} and \emph{policy improvement}. One direction for scaling the efficiency of both is to evaluate, from the current state, several possible actions, or even several possible future trajectories by using a model, instead of just extracting information from the trajectory that was executed. Those evaluations can then be used to build a locally better policy over those actions. Planning algorithms such as Monte Carlo Tree Search (MCTS) \cite{coulom:mcts} take this even further and make several local \emph{policy iteration} steps by repeatedly performing a policy improvement step followed by an explicit local step of policy evaluation of the improved policy in the aim of generating an even better policy locally.

From this perspective, the \muzero{} algorithm can be understood as the combination of two processes of \emph{policy evaluation} and \emph{policy improvement}. The inner process, concretely \muzero{}'s MCTS search, provides the \emph{policy improvement} for the outer process which in turn learns the quantities: the model, reward function, value function and the policy, necessary for the inner process. Specifically, in the outer process, \muzero{} learns a deep neural network parameterising a model, a reward function, a state-value function and a policy. \emph{Policy improvement} is accomplished by regressing the parametric policy towards the improved policy built by \muzero{}'s MCTS search. The improved policy is also used for acting. The value function is learned using the usual tools of \emph{policy evaluation} such as temporal-difference learning \cite{sutton88}. These two objectives coupled with the learning of the reward function drive the learning of the model. In the inner process, \muzero{}'s MCTS search takes several analytical \emph{policy iteration} steps: values in the search tree are estimated by explicitly averaging n-step returns bootstrapped from the value function (\emph{policy evaluation}) while visits are directed towards high policy and high value actions (\emph{policy improvement}). This results in an improved policy and an estimate of the value of this improved policy that can be used for the outer process.

This raises a few questions, especially in the case where only a small subset of the action space can be evaluated to build  the locally improved policy.
\begin{itemize}[topsep=2pt,itemsep=2pt,parsep=0pt]
\item how to select the actions or trajectories to be evaluated
\item how to build a locally improved policy over those actions
\item how to use the locally improved policy to learn about the global policy
\item how to use it to act
\item how to perform an explicit local step of policy evaluation of the improved policy for planning
\item how all these steps interact with each other
\end{itemize}

In the following, we will assume that the actions to be evaluated are sampled from some proposal distribution $\sample$ and that we have at our disposal some process to build a locally improved policy. We will mainly focus on the last four questions and propose a general framework to reason in a principled way about policy evaluation and improvement over such sampled action subsets.

\section{Sample-based Policy Iteration}
\label{sample-based-policy-improvement}
Let $\pi: \mathcal{S}\to\mathcal{P}(\mathcal{A})$ be a policy and $\ipi: \mathcal{S}\to\mathcal{P}(\mathcal{A})$ be an improved policy of $\pi$: $\forall s\in\mathcal{S}, v^{\ipi}(s) \geq v^\pi(s)$. If we had complete access to $\ipi$, we would directly use it for policy improvement by projecting it back onto the space of realisable policies. However, when the action space $\mathcal{A}$ is too large, it might only be feasible to compute an improved policy over a small subset of actions.

It is not immediately clear how to use this locally improved policy to perform principled policy improvement, or policy evaluation of the improved policy, because this locally improved policy only gives us information regarding the sampled actions.

We propose a framework which relies on writing both operations as an expectation with respect to the fully improved policy $\ipi$ and use the samples we have to estimate these expectations. This allows us to use the conceptually correct target $\ipi$ to define the objectives and clearly surface the approximations that are introduced afterwards. Specifically, we will restrict ourselves to the general class of policy improvement operators that we call \emph{action-independent} as defined below in \ref{action-indep}.

\subsection{Operator view of Policy Improvement}
\label{operator-view}
We use the concepts introduced by \cite{ghosh2020operator} and decompose \emph{policy improvement} into the successive application of two operators: (a) a policy improvement operator $\mathcal{I}$ which maps any policy to a policy achieving strictly larger return; and (b) a projection operator $\mathcal{P}$, which finds the best approximation of this improved policy in the space of realisable policies. With those notations, the process of \emph{policy improvement} can be written as $\mathcal{P}\circ\mathcal{I}$.

\cite{ghosh2020operator} showed that the policy gradient algorithm can be thought of having the following policy improvement operator: $\ipi(s, a)\propto\pi(s, a) Q(s, a)$ where $Q(s, a)$ is the action-value function. They also showed that PPO's \cite{schulman2017proximal} policy improvement operator is $\ipi(s, a)\propto\exp(Q(s, a)/\tau)$ where $\tau$ is a temperature parameter. Similarly, MPO's \cite{mpo} policy improvement operator can be written $\ipi(s, a)\propto\pi(s, a) \exp(Q(s, a)/\tau)$ and AWR \cite{peng2019advantage} uses a similar form of improved policy, replacing the action-value function by the advantage function $\ipi(s, a)\propto\pi(s, a) \exp(A(s, a)/\tau)$.

\subsection{Action-Independent Policy Improvement Operator}
\label{action-indep}
We define a policy improvement operator as \emph{action-independent} if it can be written as:
$$
\ipi(a|s) = f(s, a , Z(s))
$$
where $Z(s)$ is a unique state dependent normalising factor defined by $\forall a\in\mathcal{A}, f(s, a, Z(s))\geq 0$ and $\sum_{a\in\mathcal{A}} f(s, a, Z(s))=1$.\footnote{In the continuous case, sums would be replaced by integrals.}

All of the policy improvement operators described above are action-independent.

\textbf{MPO Example}: MPO's policy improvement operator can be written $\ipi(a|s) = f(s,a,Z(s)) = \pi(s, a) \exp(Q(s, a)/\tau)/Z(s)$ and $Z(s)=\sum_a \pi(s, a) \exp(Q(s, a)/\tau)$.

\subsection{Sample-Based Action-Independent Policy Improvement Operator}
Let $\{a_i\}$ be $K$ actions sampled from a proposal distribution $\sample$ and $\hat{\sample}(a|s)=\frac{1}{K} \sum_i \delta_{a, a_i}$ the corresponding empirical distribution\footnote{$\delta_{a,a_i}$ represents the Kronecker delta function. In the continuous case, it would be replaced by the Dirac delta function $\delta(a-a_i)$.} which is non-zero only on the sampled actions $\{a_i\}$.

We define the sample-based action-independent\footnote{We will omit the action-independent qualifier in the rest of the text when it is clear from the context.} policy improvement operator as
$$
\ibpi(a|s) = (\hat{\sample}/\sample)(a|s) f(s, a, \hat{Z}_\sample(s))
$$
where $\hat{Z}_\sample(s)$ is a unique state dependent normalising factor defined by $\forall a\in\mathcal{A}, (\hat{\sample}/\sample)(a|s) f(s, a, \hat{Z}_\sample(s))\geq 0$ and $\sum_{a\in\mathcal{A}}(\hat{\sample}/\sample)(a|s) f(s, a, \hat{Z}_\sample(s))=1$.
We have used the shorthand notation $(\hat{\sample}/\sample)(a|s)$ to mean $\hat{\sample}(a|s)/\sample(a|s)$.

\textbf{MPO Example}: MPO's sample-based action-independent policy improvement operator using $\sample=\pi$ would therefore be $\ibpi(a|s) = \hat{\sample}(a|s)\exp(Q(s, a)/\tau)/\hat{Z}_\sample(s)$ with $\hat{Z}_\sample(s)=\sum_a \hat{\sample}(a|s)\exp(Q(s, a)/\tau)$.

\subsection{Computing an expectation with respect to $\ipi$}
\label{expectation-ipi}
We focus in this section on evaluating for a given state $s$ the expectation $\mathbb{E}_{a\sim\ipi}[X|s]$ of a random variable $X$ given actions $\{a_i\}$ sampled from a distribution $\sample$ and the sample-based improved policy $\ibpi$.

\textbf{Theorem}. For a given random variable $X$, we have
$$\mathbb{E}_{a\sim\ipi}[X|s] = \lim_{K\to\infty} \sum_{a\in\mathcal{A}} \ibpi(a|s) X(s, a)$$

Furthermore, $\sum_{a\in\mathcal{A}} \ibpi(a|s) X(s, a)$ is approximately normally distributed around $\mathbb{E}_{a\sim\ipi}[X|s]$ as $K\to\infty$:
$$\sum_{a\in\mathcal{A}} \ibpi(a|s) X(s, a) \sim \mathcal{N}(\mathbb{E}_{a\sim\ipi}[X|s], \frac{\sigma^2}{K})$$
where $\sigma^2 = Var_{a\sim\beta}[\frac{f(s, a, Z(s))}{\sample} X(s, a)|s]$.

\textbf{Proof} See Appendix \ref{proofs}

\textbf{Corollary}.
The sample-based policy improvement operator converges in distribution to the true policy improvement operator:
$$\lim_{K\to\infty} \ibpi = \ipi$$ and is approximately normally distributed around the true policy improvement operator as $K\to\infty$.

\textbf{Proof}. See Appendix \ref{proofs}

\begin{figure*}
\includegraphics[trim=0 0 0 4,clip,width=\textwidth]{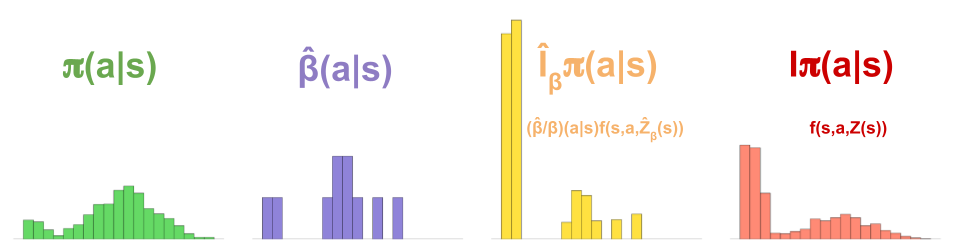}
\vspace*{-7mm}
\caption[]{
\label{fig:sample-policy-improvement}
\textbf{Sample-based Policy Improvement.} On the left, the current policy $\pi(a|s)$. Next, $K$ actions $\{a_i\}$ are sampled from a proposal distribution $\sample$ and $\hat{\sample}(a|s)$ is the corresponding empirical distribution. A sample-based improved policy $\ibpi(a|s)= (\hat{\sample}/\sample)(a|s) f(s, a, \hat{Z}_\sample(s))$ is then built. As the number of samples $K$ increases  $\ibpi(a|s)$ converges to the improved policy $\ipi(a|s)=f(s, a, Z(s))$.
}
\end{figure*}

We illustrate this result in Figure \ref{fig:sample-policy-improvement}.

\subsection{Sample-based Policy Evaluation and Improvement}
The previous expression computing an estimate of $\mathbb{E}_{a\sim\ipi}[X|s]$ using the quantity $\ibpi$ and the sampled actions $\{a_i\}$ can be used for policy improvement and policy evaluation of the improved policy.

Policy improvement can be performed by for instance instantiating $X=-\log\pi_\theta$, minimising the cross-entropy between $\pi_\theta$ and the improved policy $\ipi$: $CE=\mathbb{E}_{a\sim\ipi}[-\log\pi_\theta]$.

Additionally, samples from $\ipi$ can be obtained by re-sampling an action $a$ from $\ibpi$. This procedure also known as Sampling Importance Resampling (SIR) \cite{sir} gives us a way to act with the improved policy and reuse the usual tools such as temporal-difference learning to do policy evaluation of the improved policy.

Finally, for instance for the purpose of planning, an explicit step of policy evaluation of the improved policy can be computed by estimating 1-step or n-step returns. Using for example $X=r +\gamma V'$ lets us backpropagate the value $V'$ by one step in a search tree: $V(s)=\mathbb{E}_{a\sim\ipi}[r+\gamma V'|s]$.

\section{\smuzero{}}
Building on the sample-based policy iteration framework established in the previous section, we now instantiate those ideas in the context of a complete system. Concretely, we apply our sampling procedure to the \muzero{} algorithm, to produce a new algorithm that we term \smuzero{}. This algorithm may be applied to any domain where \muzero{} can be applied; but furthermore can also be used, in principle, to learn and plan in domains with arbitrarily complex action spaces.

As introduced in the background section, \muzero{} may be understood as combining an inner process of policy iteration, within its Monte-Carlo tree search, and an outer process, in its overall interactions with the environment.

\subsection{Inner Policy Evaluation and Improvement}
Specifically, within its search tree, \muzero{} estimates values by explicitly averaging n-step returns samples (\emph{policy evaluation}) and selects the next node to evaluate and expand by recursively maximising (\emph{policy improvement}) over the probabilistic upper confidence tree (PUCT) bound \cite{Silver16AG} $$arg\,max_a Q(s,a) + c(s) \cdot \pi(s,a)\frac{\sqrt{\sum_b N(s,b)}}{1+N(s,a)}$$ where $c(s)$ is an exploration factor controlling the influence of the policy $\pi(s,a)$ relative to the values $Q(s,a)$ as nodes are visited more often.

\textbf{Naive Modification.}
A first approach to extending \muzero{}'s MCTS search is to search over the sampled actions $\{a_i\}$ and keep the PUCT formula unchanged, directly using the probabilities $\pi$ coming from the policy network in the PUCT formula just like in \muzero{}. The search's visit count distribution $f$ can then be used to construct the sampled-based $\ibpi = \hat{\sample}/\sample f$ to correct for the effect of sampling at policy network training time and acting time, but also dynamically as the tree is built for value backpropagation (inner \emph{policy evaluation}). Theoretically this procedure is not complicated, but in practice it might lead to unstable results because of the $f/\sample$ term, especially if $f$ is represented by normalised visit counts which have limited numerical precision.

\textbf{Proposed Modification.}
Instead, we propose to search with probabilities $\hat{\pi}_\sample(s, a)$ proportional to $(\hat{\sample}/\sample \pi)(s,a)$, in place of $\pi(s, a)$ in the PUCT formula and directly use the resulting visit count distributions just like in \muzero{}. We use the following Theorem to justify this proposed modification.

\textbf{Theorem}.
Let $\ipi$ be the visit count distribution\footnote{for a given number of simulations} of \muzero{}'s search using prior $\pi$ when considering the whole action space $\mathcal{A}$ and let $\ihpi$ be the visit count distribution obtained by searching using prior $\hat{\pi}_\sample$. Then, $\ihpi$ is approximately equal to the sample-based policy improvement associated to $\ipi$. In other words, $\ihpi \approx \ibpi$.

\textbf{Proof}. See Appendix \ref{mz-pi}

We can therefore directly use the results of the previous section \ref{expectation-ipi} and in particular,
$$
\mathbb{E}_{a\sim\ipi}[X|s] \approx \sum_{a\in\mathcal{A}}\ihpi(s, a) X(s,a)
$$
This lets us conclude that the only modification beyond sampling that needs to be made to \muzero{} is to use $\hat{\pi}_\sample$ instead of $\pi$ in the PUCT formula. The rest of the \muzero{} algorithm, from estimating the values in the search tree by averaging n-step returns, to acting and training the policy network using the visit count distribution, can proceed unchanged.

\textbf{Remark.} Note that, if $\sample=\pi^{1/\tau}$, the probabilities $\hat{\pi}_\sample$ used in the PUCT formula can be written: $\hat{\pi}_\sample=\hat{\sample}/\sample \pi= \hat{\sample} \pi^{1-1/\tau}$. If $\tau=1$, $\hat{\pi}_\sample$ is equal to the empirical sampling/prior distribution $\hat{\sample}$. This means that the search is guided by a potentially quasi uniform prior $\hat{\sample}$ but only evaluates relatively high probability actions. If $\tau>1$, the search evaluates more diverse samples but is guided by more peaked probabilities $\hat{\sample} \pi^{1-1/\tau}$.

\subsection{Outer Policy Improvement}

Once the inner iterations of policy improvement and policy evaluation within Monte-Carlo tree search have been completed, the net result is a set of visit counts $N(s,a)$ at the root state $s$ of the search tree, corresponding to each sampled action $a$. These visit counts may be normalised to provide the sample-based improved policy $\ihpi(a|s) = N(s,a) / \sum_b N(s,b)$. Following the argument in the previous section, these visit counts already take account of the fact that the root actions were sampled according to $\sample$.

Hence all that remains is to project the sample-based improved policy back onto the space of representable policies, using an appropriate projection operator $\mathcal{P}$. Following \muzero{}, we choose a standard projection operator for probability distributions that selects parameters $\theta$ minimising the KL divergence $KL(\ihpi || \pi_\theta)$.

\subsection{Outer Policy Evaluation}

To select actions, the agent samples its behaviour from its sample-based improved policy, $\ihpi(a|s) = N(s,a) / \sum_b N(s,b)$. As above, we note that this already corrects for the sampling procedure in the construction of the visit counts, and hence may be used directly as a policy.

The outer policy evaluation step then follows directly from \muzero{}, i.e. a value function is trained from $n$-step returns, using trajectories of behaviour generated by the sample-based improved policy.

\subsection{Search Tree Node Expansion}

In \muzero{}, each time a leaf node is expanded, all the $N=|\mathcal{A}|$ actions of the action space are returned alongside the probabilities $\pi$ the policy network assigns to each of those actions.

\textbf{Proposed Modification.}
In \smuzero{}, we instead sample $K\ll N$ actions from a distribution $\sample$ and return each action $a$ along with its corresponding probabilities $\pi(s,a)$ and $\sample(s,a)$.%

We note that, if the number of simulations of the search is much bigger than $K$, techniques such as progressive widening \cite{progressive_widening} could in principle be used to dynamically sample more actions for nodes on highly visited search paths.

\subsection{Sampling distribution $\sample$}

In principle any sampling distribution $\sample$ with a wide support can be used, including the uniform distribution. However, as only a limited number of samples can be drawn, it is preferable to sample moves that are likely according to our current estimate for the best policy, i.e. the policy network.\footnote{Note that \muzero{} with a limited number of simulations will only visit the high prior moves}

\textbf{Proposed Modification.}
We use $\sample=\pi$, potentially modulated by a temperature parameter. To encourage exploration and to make sure that even low prior moves have an opportunity to be reassessed from time to time, \muzero{} combines the prior $\pi$ produced by the policy network with Dirichlet noise at the root of the search tree. We obtain the same behaviour in \smuzero{} by also including noise in $\sample$ and $\pi$, ensuring that low prior moves can be sampled and searched.

\section{Experiments}

We evaluated the performance of \smuzero{} on a variety of reinforcement learning environments. We focus upon standard benchmark environments in which clear baselines are available for comparison. We use those benchmarks to explore two important properties of real-world applications. First, whether \smuzero{} is sufficiently general to operate across discrete and continuous environments of very different types. Second, whether the algorithm is robust to sampling -- that is, whether we can come close to the performance of algorithms that have access to the entire action set (and are therefore not scalable to large action spaces), when only sampling a small fraction of the action space.

\subsection{Go}

Go has long been a challenge problem for AI, with only the \emph{AlphaGo} \cite{Silver16AG,Silver18AZ,muzero} family of algorithms finally surpassing human professional players. It is a domain that requires deep and precise planning and as such is an ideal domain to put the planning capabilities of \smuzero{} to the test.

\begin{figure}
\includegraphics[width=\columnwidth]{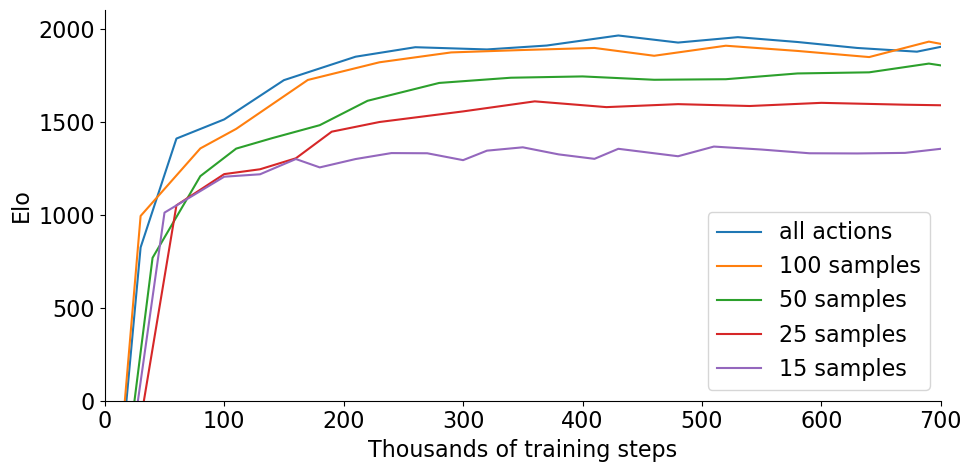}
\vspace*{-7mm}
\caption[]{
\label{fig:go-results}
\textbf{Results in the classical board game of Go.} Performance of \smuzero{} (1 seed per experiment) with different number $K \in \{15, 25, 50, 100\}$ of samples throughout training compared to a \muzero{} baseline that always considers all available actions (action space of size 362). Elo scale anchored to final baseline performance of 2000 Elo. As the number of sampled actions increases, performance rapidly approaches the baseline that always considers all actions.
}
\end{figure}

Using \muzero{} as a baseline, we trained multiple instances of \smuzero{} with varying number of action samples $K \in \{15, 25, 50, 100\}$ (see Figure \ref{fig:go-results}). The size of the action space in 19x19 Go is 362 (all board points plus pass), so all the tested values of $K$ only cover a small part of the action space. As expected, the performance improves as $K$ increases, with $K=50$ samples already closely approaching the performance of the baseline that is allowed to search over all possible actions.

\subsection{Atari}
\begin{figure}
\includegraphics[width=\columnwidth]{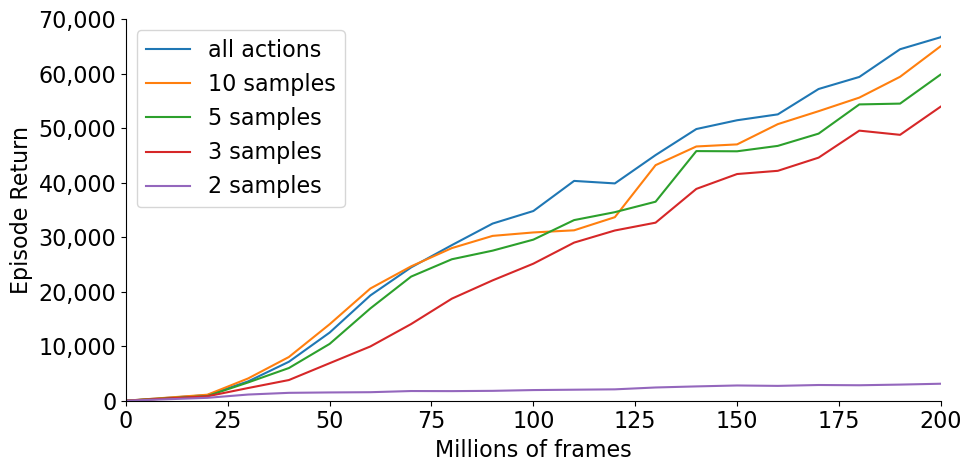}
\vspace*{-7mm}
\caption[]{
\label{fig:pacman-results}
\textbf{Results in Ms. Pacman.} Performance of \smuzero{} (1 seed per experiment) with different number of samples throughout training compared to a \muzero{} baseline that always considers all available actions. As the number of sampled actions increases, performance rapidly approaches the baseline that always considers all actions.
}
\end{figure}

We also performed the same experiment as in Figure \ref{fig:go-results} for the Arcade game of Ms. Pacman, from the classic Atari RL benchmark. The action space in Atari is of size 18. Searching with $K = 2$ samples is not sufficient for efficient learning, but already with $K = 3$ samples performance rapidly approaches the baseline that is allowed to search all possible actions without sampling (Figure \ref{fig:pacman-results}).

\subsection{\dmcs}

\begin{figure*}
\includegraphics[width=\textwidth]{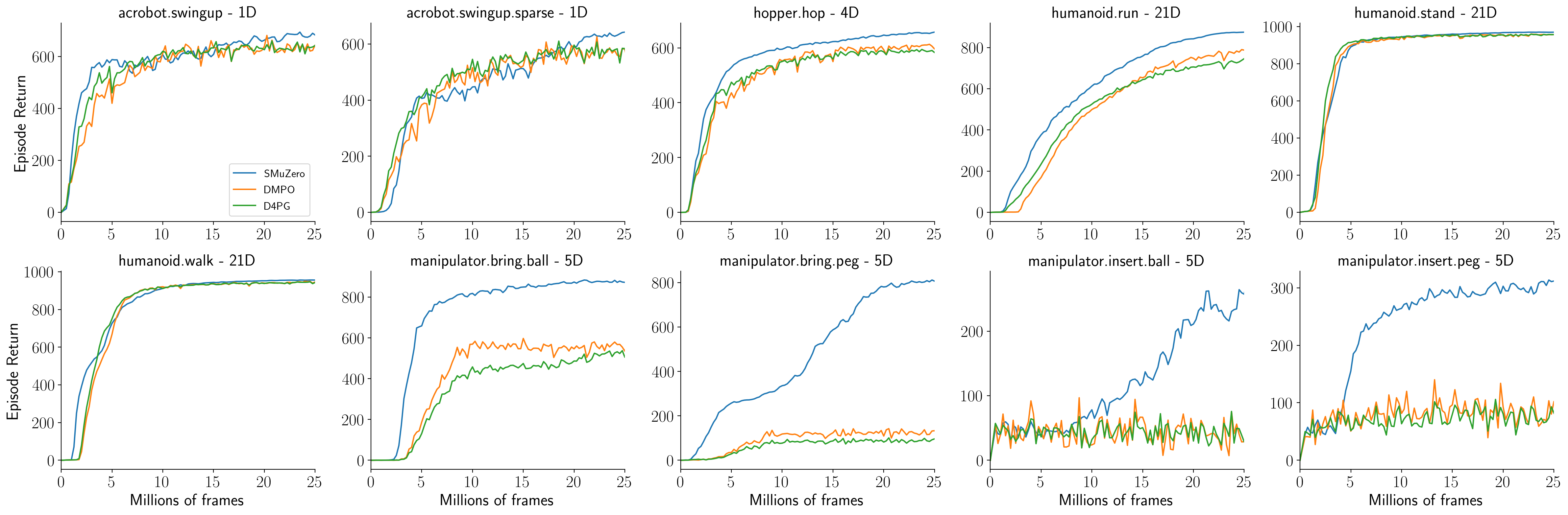}
\vspace*{-7mm}
\caption[]{
\label{fig:control-suite-hard-results}
\textbf{Results in DM Control Suite Hard and Manipulator tasks.} Performance of \smuzero{} (3 seeds per experiment) throughout training compared to DMPO \cite{hoffman2020acme} and D4PG \cite{d4pg}. The x-axis shows millions of environment frames, the y-axis mean episode return. Hard tasks as proposed by \cite{hoffman2020acme}. Plot titles include the task name and the dimensionality of the action space.
}
\end{figure*}

\begin{figure*}
\includegraphics[width=\textwidth]{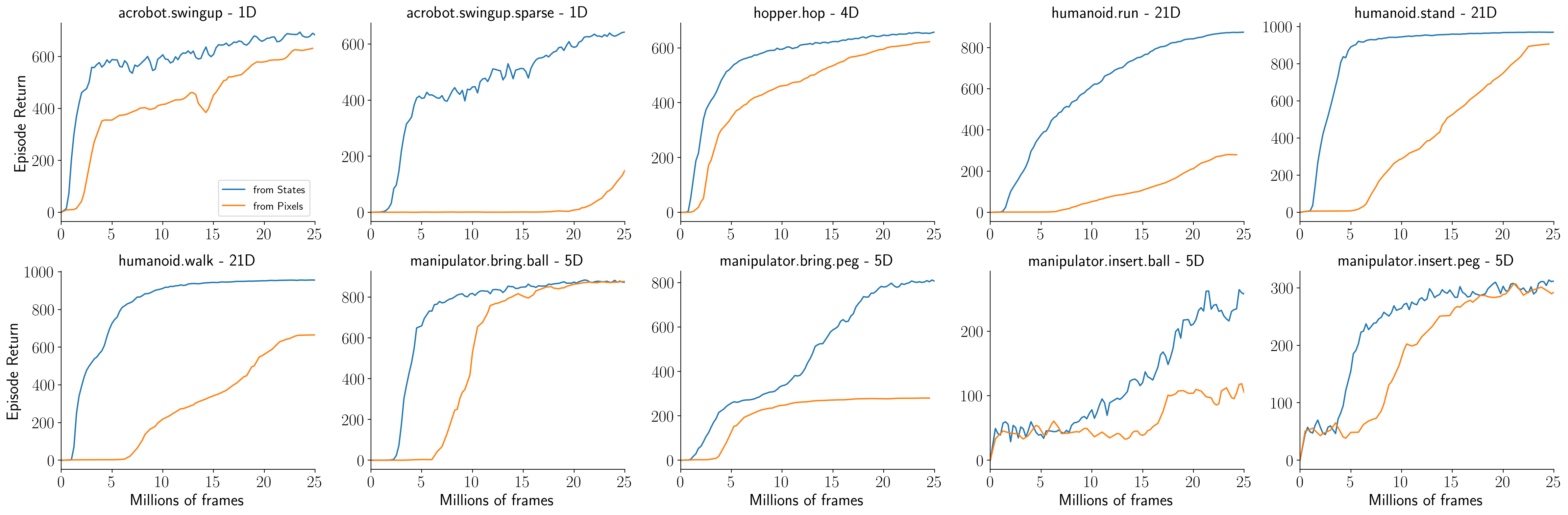}
\vspace*{-7mm}
\caption[]{
\label{fig:control-suite-pixel-results}
\textbf{Results in DM Control Suite Hard and Manipulator tasks of \smuzero{} learning from raw pixel inputs.} Performance of \smuzero{} (3 seeds per experiment) learning from raw pixel inputs throughout training compared to \smuzero{} learning from state inputs. The x-axis shows millions of environment frames, the y-axis mean episode return.
}
\end{figure*}

\begin{figure*}
\includegraphics[trim=0 0 320 0,clip,width=\textwidth]{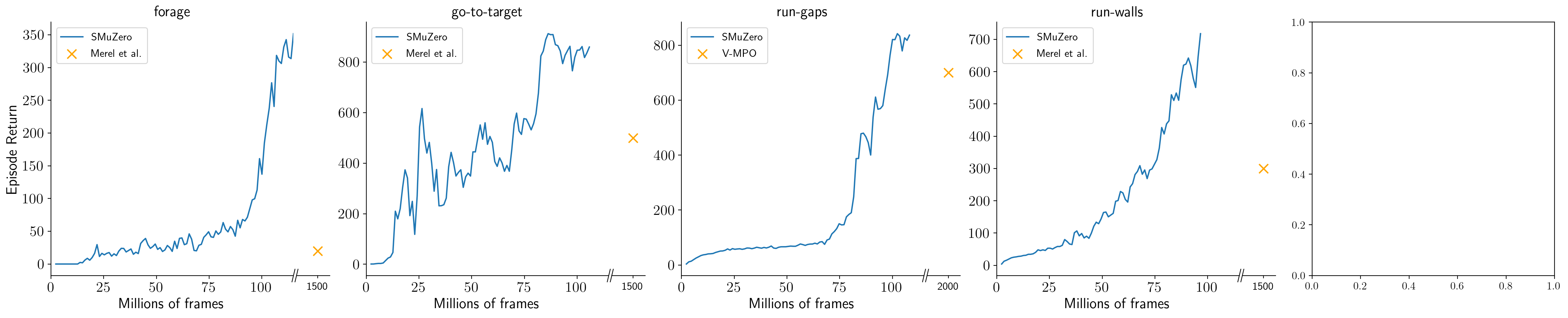}
\vspace*{-7mm}
\caption[]{
\label{fig:cmu-humanoid-results}
\textbf{Results for 56D CMU Humanoid Locomotion tasks.} Performance of \smuzero{} (1 seed per experiment) throughout training in dm\_control \cite{tassa2020dm_control} based CMU humanoid tasks, controlling a humanoid body with 56 action dimensions. \smuzero{} outperforms previously reported results for both forage, go-to-target and run-walls \cite{merel2018hierarchical} as well as run-gaps \cite{Song2020VMPO} while using more than an order of magnitude fewer environment interactions.
}
\end{figure*}

The \dmcs \cite{tassa2018deepmind} provides a set of continuous control tasks based on MuJoCo \cite{mujoco} and has been widely used as a benchmark to assess performance of continuous control algorithms. For the experiments in this paper we use the task classification and data budgets introduced in Acme \cite{hoffman2020acme}, evaluating \smuzero{} on the easy, medium and hard tasks. We additionally evaluated \smuzero{} on the manipulator tasks which are known to be interesting and difficult.

In its most common setup, the control suite domains provide 1 dimensional state inputs (as opposed to 2 dimensional image inputs in board games and Atari as used by \muzero{}). We therefore used a variation of the \muzero{} model architecture in which all convolutions are replaced by fully-connected layers (see Appendix \ref{cs-experiments} for further details). For the policy prediction, we chose the factored policy representation introduced by \cite{tang2020discretizing}, representing each dimension by a categorical distribution. There are however no difficulties in working directly with continuous actions and we show results with a policy prediction parameterised with a Gaussian distribution on the hard and manipulator tasks in the Appendix (Figure \ref{gaussian}).

\smuzero{} showed good performance across the task set (Figure \ref{fig:control-suite-results} for full results), with especially good results for tasks in the most difficult hard and manipulator categories (Figure \ref{fig:control-suite-hard-results}) such as \emph{humanoid.run} or the \emph{manipulator} tasks in general.

The control suite domains can also be configured to provide raw pixel inputs instead of 1 dimensional state inputs. We ran \smuzero{} on the same tasks with the same data budget (25M frames) and the same hyperparameters. As demonstrated in Figure \ref{fig:control-suite-pixel-results}, \smuzero{} can be applied to efficiently learn from raw pixel inputs as well. It is particularly remarkable that \smuzero{} can learn to control the 21 dimensional humanoid from raw pixel inputs only. In addition, we compared \smuzero{} to the Dreamer agent \cite{hafner2019dream} in Appendix \ref{dreamer}, Table \ref{tab:dreamer-results}. \smuzero{} equalled or surpassed the Dreamer agent's performance in all tasks, without any action repeat (Dreamer uses an action repeat of 2), observation reconstruction, or any hyperparameter re-tuning.

To investigate the scalability to more complex action spaces, we also applied \smuzero{} to the dm\_control \cite{tassa2020dm_control} based Locomotion environment. In this set of high-dimensional tasks, the agent must control a humanoid body with 56 action dimensions to accomplish a variety of goals (Figure \ref{fig:cmu-humanoid-results}). In all tasks \smuzero{} not only outperformed previously reported results, but it did so using more than an order of magnitude fewer interactions with the environment.

Finally, we investigated the impact on performance of the number of samples in the Appendix (Figure \ref{fig:control-suite-samples}). We show that \smuzero{} can learn high dimensional action tasks with as little as $K=5$ samples. Furthermore, we evaluated the stability of \smuzero{}, both from state inputs and raw pixel inputs, in Figure \ref{fig:hard-results-seeds} and Figure \ref{fig:control-suite-reproducibility}. We show that \smuzero{}'s performance is overall very reproducible across tasks and number of samples. We also verified the practical importance of using $\hat{\pi}_\sample$ instead of just $\pi$ in \smuzero{}'s PUCT formula in Figure \ref{fig:control-suite-ptilde}. We find that, as suggested by the theory, it is much more robust to use $\hat{\pi}_\sample$.

\subsection{\rwrl Challenge Benchmark}

The real-world Reinforcement Learning (RWRL) Challenge set of benchmark tasks \cite{dulacarnold2020empirical} is a set of continuous control tasks that aims to capture the aspects of real-world tasks that commonly cause RL algorithms to fail. We used this benchmark to test the robustness of our proposed algorithm to complications such as delays, partial observability or stochasticity. We used the same neural network architecture as for the \dmcs with the addition of an LSTM \cite{lstm} to deal with partial observability.

As shown in Table \ref{tab:rwrl-results}, \smuzero{} significantly outperformed baseline algorithms in all three challenge difficulties. We provide full learning curve results in the Appendix (Figure \ref{fig:rwrl-results}).

\begin{table}[t]
\begin{tabularx}{\columnwidth}{l rrrr}

\toprule
Agent & Cartpole & Walker & Quadruped & Humanoid \\\midrule
\multicolumn{5}{c}{Easy} \\
DMPO & 464.05 & 474.44 & 567.53 & 1.33\\
D4PG & 482.32 & 512.44 & 787.73 & 102.92\\
STACX & 734.40 & 487.75 & 865.80 & 1.21\\
SMuZero & \textbf{861.05} & \textbf{959.83} & \textbf{987.20} & \textbf{289.36}\\
\midrule
\multicolumn{5}{c}{Medium} \\
DMPO & 155.63 & 64.63 & 180.30 & 1.27\\
D4PG & 175.47 & 75.49 & 268.01 & 1.28\\
STACX & 398.71 & 94.01 & 466.43 & 1.18\\
SMuZero & \textbf{516.69} & \textbf{448.51} & \textbf{946.21} & \textbf{108.56}\\
\midrule
\multicolumn{5}{c}{Hard} \\
DMPO & 138.06 & 63.05 & 144.69 & 1.40\\
D4PG & 108.20 & 59.85 & 280.75 & 1.27\\
STACX & 135.26 & 58.11 & \textbf{351.56} & 1.26\\
SMuZero & \textbf{244.71} & \textbf{71.16} & 348.09 & 1.19\\
\bottomrule
\end{tabularx}

\caption{
\label{tab:rwrl-results}
\textbf{\smuzero{} results for the Real-Word RL benchmark} (RWRL). In RWRL, for each task there is an easy, medium and hard variation of increasing difficulty. DMPO and D4PG results from \cite{dulacarnold2020empirical}, STACX from \cite{zahavy2020stacx}. \smuzero{} (3 seeds per experiment) shows strong performance throughout, especially on the highest dimensional Humanoid tasks.
}
\end{table}

\section{Conclusions}

In this paper we introduced a unified framework for learning and planning in discrete, continuous and structured complex action spaces. Our approach is based upon a simple principle of sampling actions. By careful book-keeping we have shown how one may take account of the sampling process during policy improvement and policy evaluation. In principle, the same sample-based strategy could be applied to a variety of other reinforcement algorithms in which the policy is updated by, or approximated by, an action-independent improvement step. Concretely, we have focused upon applying our framework to the model-based planning algorithm of \muzero{}, resulting in our new algorithm \smuzero{}. Our empirical results show that the idea is both general, succeeding across a wide variety of discrete and continuous benchmark environments, and robust, scaling gracefully down to small numbers of samples. These results suggest that the ideas introduced in this paper may also be effective in larger scale applications where it is not feasible to enumerate the action space.

\section*{Acknowledgements}
We would like to thank Jost Tobias Springenberg for providing very detailed feedback and constructive suggestions.

\bibliography{main}
\bibliographystyle{icml2020}

\clearpage

\appendix

\section{\dmcs and \rwrl Experiments}
\label{cs-experiments}
For the continuous control experiments where the input is 1 dimensional (as opposed to 2 dimensional image inputs in board games and Atari as used by \muzero{}), we used a variation of the \muzero{} model architecture in which all convolutions are replaced by fully connected layers.

The representation function processed the input via an \emph{input block} composed of a linear layer, followed by a Layer Normalisation and a \emph{tanh} activation. The resulting embedding was then processed by a ResNet v2 style pre-activation residual tower \cite{resv2} coupled with Layer Normalisation \cite{ba2016layer} and Rectified Linear Unit (ReLU) activations. We used 10 blocks, each block containing 2 layers with a hidden size of 512.

For the \rwrl experiments, we additionally inserted an LSTM module \cite{lstm} in the representation function between the \emph{input block} and the residual tower to deal with partial observability. We trained the LSTM using truncated backpropagation through time for 8 steps, initialised from LSTM states stored during acting, each step having the last $4$ observations concatenated together, for an effective unroll step of $32$ steps.

The dynamics function processed the action via an \emph{action block} composed of a linear layer, followed by a Layer Normalisation and a \emph{ReLU} activation. The action embedding was then added to the dynamics function's input embedding and then processed by a residual tower using the same architecture as the residual tower for the representation function.

The reward and value predictions used the categorical representation introduced in \muzero{} \cite{muzero}. We used 51 bins for both the value and the reward predictions with the value being able to represent values between $\left[-150.0, 150.0\right]$ and the reward being able to represent values between $\left[-1.0, 1.0\right]$. We used n-step bootstrapping with $n=5$ and a discount of $0.99$ consistent with Acme \cite{hoffman2020acme}.

We used the factored policy representation introduced by \cite{tang2020discretizing} representing each dimension by a categorical distribution over $B=7$ bins for the policy prediction.

To implement the network, we used the modules provided by the Haiku neural network library \cite{haiku2020github}.

We used the Adam optimiser \cite{adam} with decoupled weight decay \cite{adam_weight_decay} for training. We used a weight decay scale of $2 \cdot{} 10^{-5}$, a batch size of $1024$ an initial learning rate of $10^{-4}$, decayed to 0 over 1 million training batches using a cosine schedule:

$$\text{lr} = \text{lr}_\text{init} \frac{1}{2} \left(1+ \cos{\pi \frac{\text{step}}{\text{max\_steps}} }\right)$$

where $\text{lr}_\text{init} = 10^{-4}$ and $\text{max\_steps} = 10^6$.

For replay, we keep a buffer of the most recent $2000$ sequences, splitting episodes into subsequences of length up to $500$. Samples are drawn from the replay buffer according to prioritised replay \cite{Schaul2016} using the same priority and hyperparameters as in \muzero{}.

We trained \smuzero{} using $K=20$ samples and a search budget of $50$ simulations per move. At the root of the search tree only, we evaluated all sampled actions before the start of the search and used those to initialise the $Q(s,a)$ quantities in the PUCT formula (Appendix \ref{search}). We evaluated \smuzero{}'s network checkpoints throughout training playing $100$ games with a search budget of $50$ simulations per move and picked the move with the highest number of visits to act, consistent with previous work.

We used Acme \cite{hoffman2020acme} to produce the results for DMPO \cite{hoffman2020acme} and D4PG \cite{d4pg}. Compared to Acme, we used bigger networks (Policy Network layers = (512, 512, 256, 128), Critic Network Layers = (1024, 1024, 512, 256)) and a bigger batch size of $1024$ for better comparison. Each task was run with three seeds.

\begin{figure*}
\includegraphics[width=\textwidth]{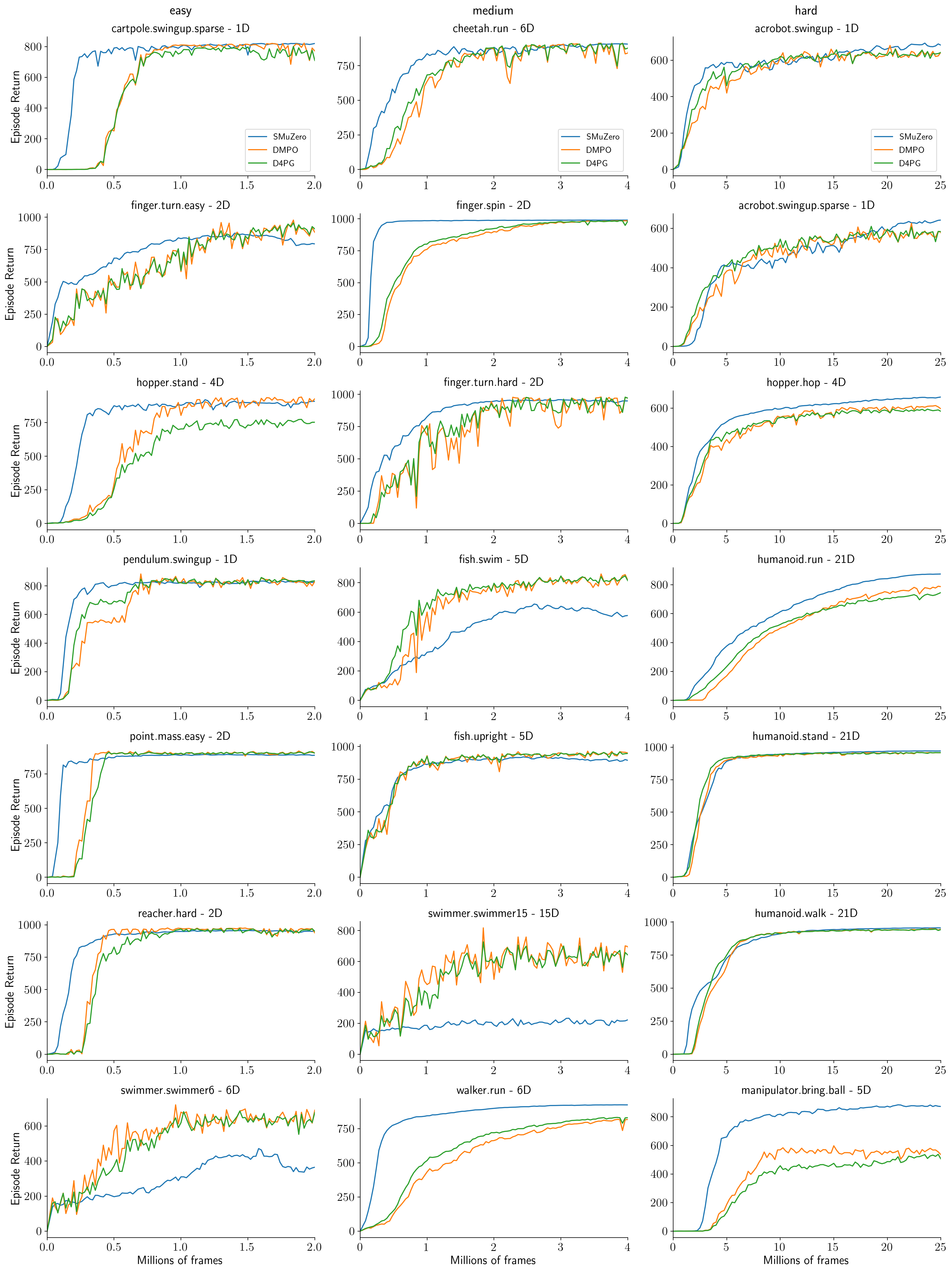}
\vspace*{-7mm}
\caption[]{
\label{fig:control-suite-results}
\textbf{Results in DM Control Suite tasks.} Performance of \smuzero{} (3 seeds per experiment) throughout training compared to DMPO \cite{hoffman2020acme} and D4PG \cite{d4pg}. The x-axis shows millions of environment frames, the y-axis mean episode return. Tasks are grouped into easy, medium and hard as proposed by \cite{hoffman2020acme}. Plot titles include the task name and the dimensionality of the action space.
}
\end{figure*}

\begin{figure*}
\includegraphics[width=\textwidth]{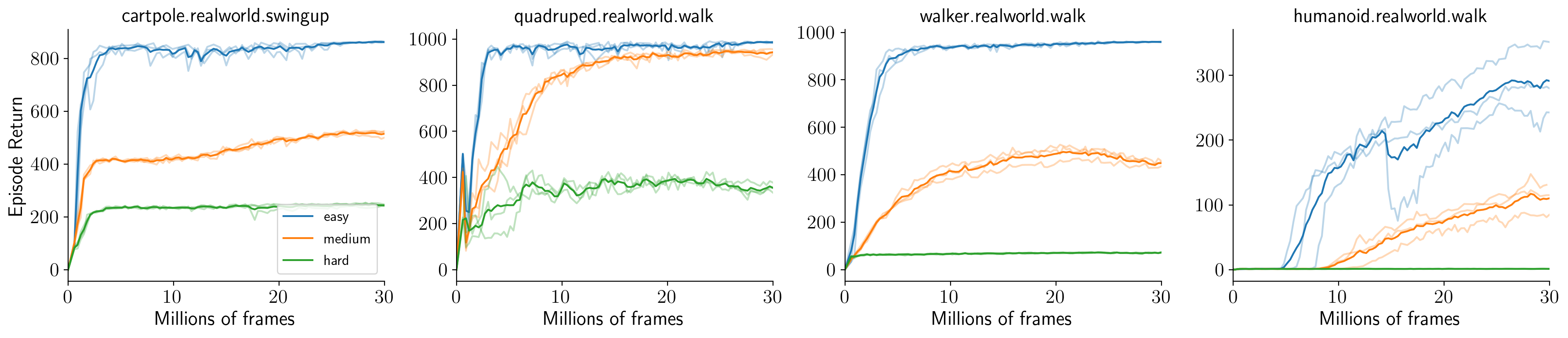}
\vspace*{-7mm}
\caption[]{
\label{fig:rwrl-results}
\textbf{\smuzero{} results for the Real-Word RL benchmark.} Performance of \smuzero{} (3 seeds per experiment) throughout training on the easy, medium and hard variations of difficulty. The x-axis shows millions of environment frames, the y-axis mean episode return. Tasks are grouped into easy, medium and hard. Plot titles include the task name.
}
\end{figure*}

We provide full learning curve results on the \dmcs (Figure \ref{fig:control-suite-results}) and \rwrl (Figure \ref{fig:rwrl-results}) tasks.

\subsection{Gaussian policy parameterisation}
\label{gaussian}
\begin{figure*}
\includegraphics[width=\textwidth]{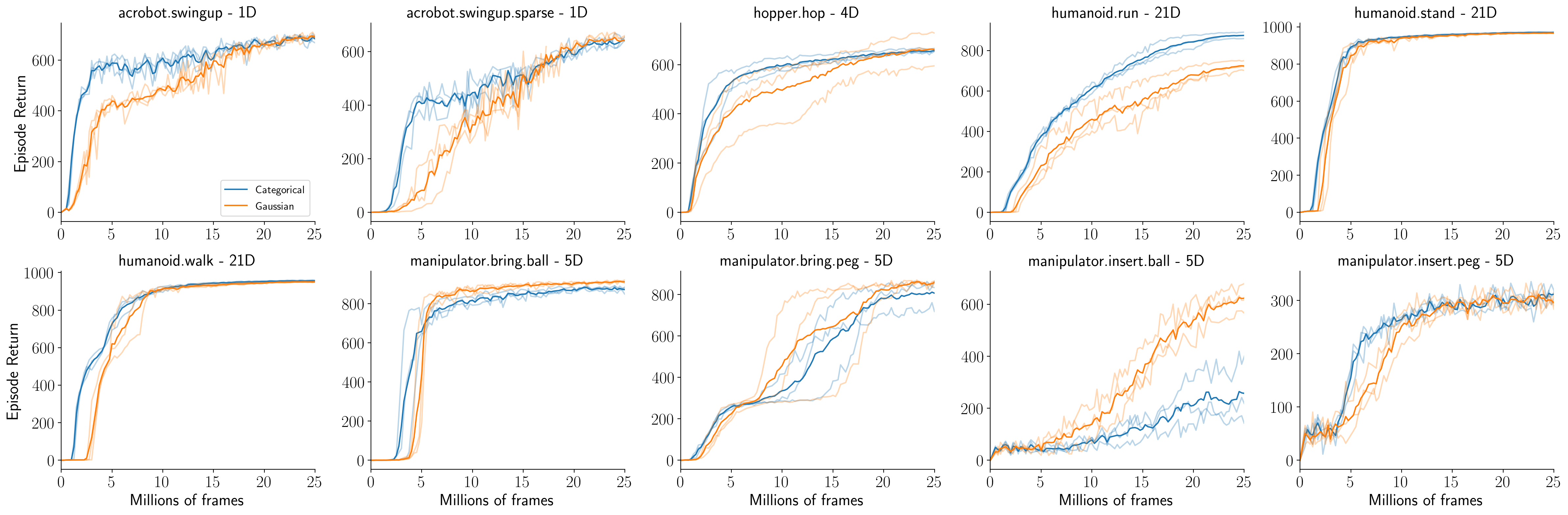}
\vspace*{-7mm}
\caption[]{
\label{fig:control-suite-gaussian-results}
\textbf{Comparison between a Categorical and Gaussian parameterisation of the policy prediction for \smuzero{}.} Performance of \smuzero{} (3 seeds per experiment) throughout training on the DM Control Hard and Manipulator tasks.
}
\end{figure*}

Even though a categorical policy representation was used to compute the main results, \smuzero{} can also be applied working directly with continuous actions. Figure \ref{fig:control-suite-gaussian-results} shows results on the hard and manipulator tasks when the policy prediction is parameterised by a Gaussian distribution.

The performance is similar across almost all tasks but we found that Gaussian distributions are harder to optimise than their categorical counterpart and  that using entropy regularisation was useful to produce better results (we used a coefficient of 5e-3). It is possible that these results could be improved with better regularisation schemes such as constraining the deviation of the mean and standard deviation as in the MPO \cite{mpo} algorithm. In contrast, we did not need to add any regularisation to train the categorical distribution.

\subsection{\smuzero{} from Pixels}
\label{dreamer}
In addition to \smuzero{}'s results on the hard and manipulator tasks when learning from raw pixel inputs, we compared \smuzero{} to the Dreamer agent \cite{hafner2019dream} in Table \ref{tab:dreamer-results}. We used the 20 tasks and the 5 million environment steps experimental setup defined by \cite{hafner2019dream}. \smuzero{} equalled or surpassed the Dreamer agent's performance in all 20 tasks, without any action repeat (Dreamer uses an action repeat of 2), observation reconstruction, or any hyperparameter re-tuning.

\begin{table}[t]
\begin{tabularx}{\columnwidth}{l rr}

\toprule
Tasks & Dreamer & SMuZero \\
\midrule
acrobot.swingup & 365.26 & \textbf{417.52}\\
cartpole.balance & \textbf{979.56} & \textbf{984.86}\\
cartpole.balance\_sparse & 941.84 & \textbf{998.14}\\
cartpole.swingup & 833.66 & \textbf{868.87}\\
cartpole.swingup\_sparse & 812.22 & \textbf{846.91}\\
cheetah.run & 894.56 & \textbf{914.39}\\
ball\_in\_cup.catch & 962.48 & \textbf{977.38}\\
finger.spin & 498.88 & \textbf{986.38}\\
finger.turn\_easy & 825.86 & \textbf{972.53}\\
finger.turn\_hard & 891.38 & \textbf{963.07}\\
hopper.hop & 368.97 & \textbf{528.24}\\
hopper.stand & \textbf{923.72} & \textbf{926.50}\\
pendulum.swingup & \textbf{833.00} & \textbf{837.76}\\
quadruped.run & 888.39 & \textbf{923.54}\\
quadruped.walk & \textbf{931.61} & \textbf{933.77}\\
reacher.easy & 935.08 & \textbf{982.26}\\
reacher.hard & 817.05 & \textbf{971.53}\\
walker.run & 824.67 & \textbf{931.06}\\
walker.stand & \textbf{977.99} & \textbf{987.79}\\
walker.walk & 961.67 & \textbf{975.46}\\
\bottomrule
\end{tabularx}

\caption{
\label{tab:dreamer-results}
\textbf{Performance of \smuzero{} compared to the Dreamer agent.} \smuzero{} equals or outperforms the Dreamer agent in all tasks. Dreamer results from \cite{hafner2019dream}.
}
\end{table}

\subsection{Ablation on the number of samples}
\begin{figure}
\includegraphics[width=\columnwidth]{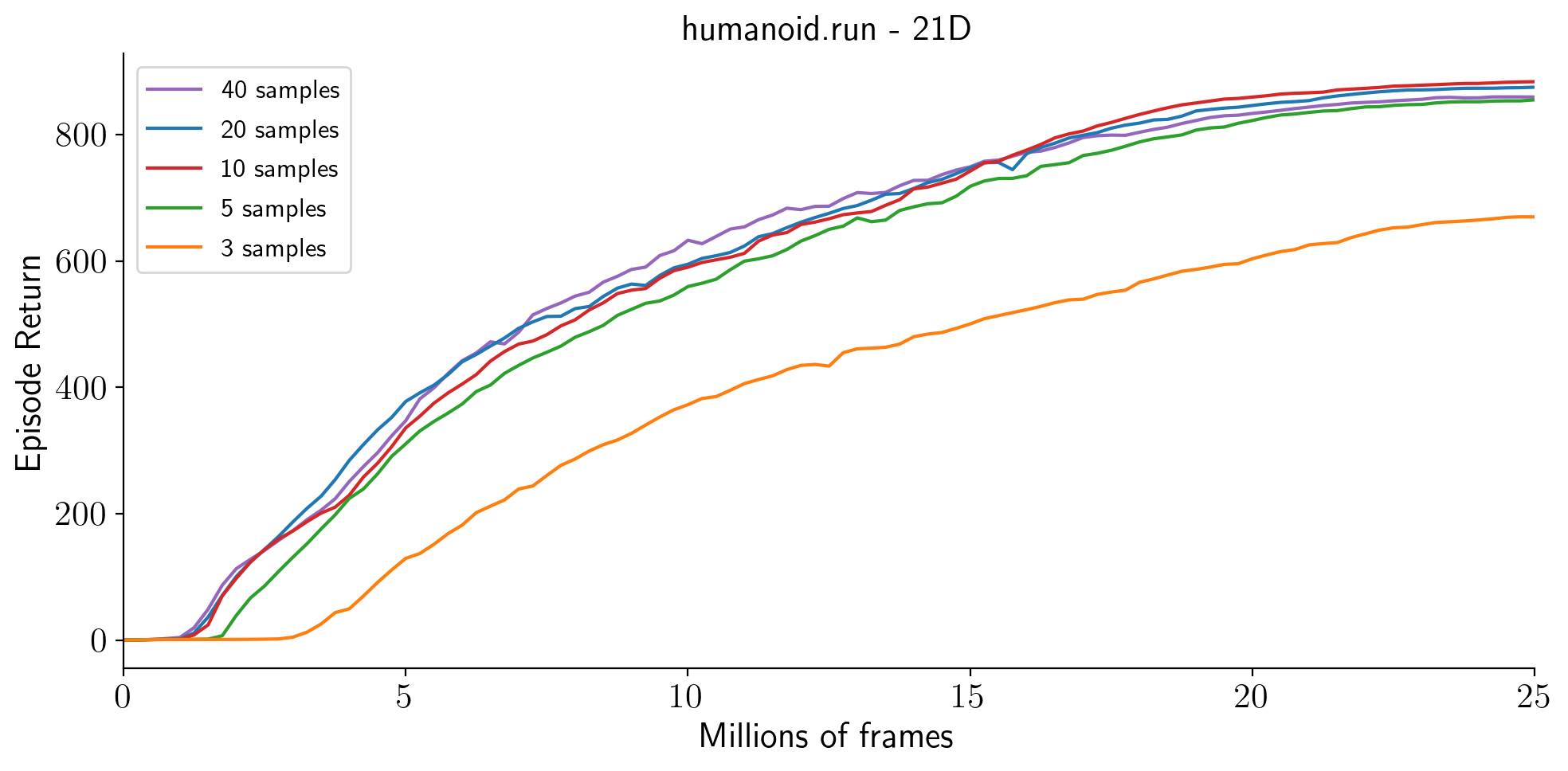}
\vspace*{-7mm}
\caption[]{
\label{fig:control-suite-samples}
\textbf{Performance of \smuzero{} with different number of samples on the \emph{humanoid.run} task.} Performance of \smuzero{}  (6 seeds per experiment) throughout training on the DM Control Humanoid Run task.
}
\end{figure}

We trained multiple instances of \smuzero{} with varying number of action samples $K \in \{3,5,10,20,40\}$ on the \emph{humanoid.run} task for which the action is 21 dimensional. We ran six seeds for each instance. Surprisingly $K=3$ is already sufficient to learn a good policy and performance does not seem to be improved by sampling more than $K=10$ samples (see Figure \ref{fig:control-suite-samples}).

\subsection{Reproducibility}
\begin{figure*}
\includegraphics[width=\textwidth]{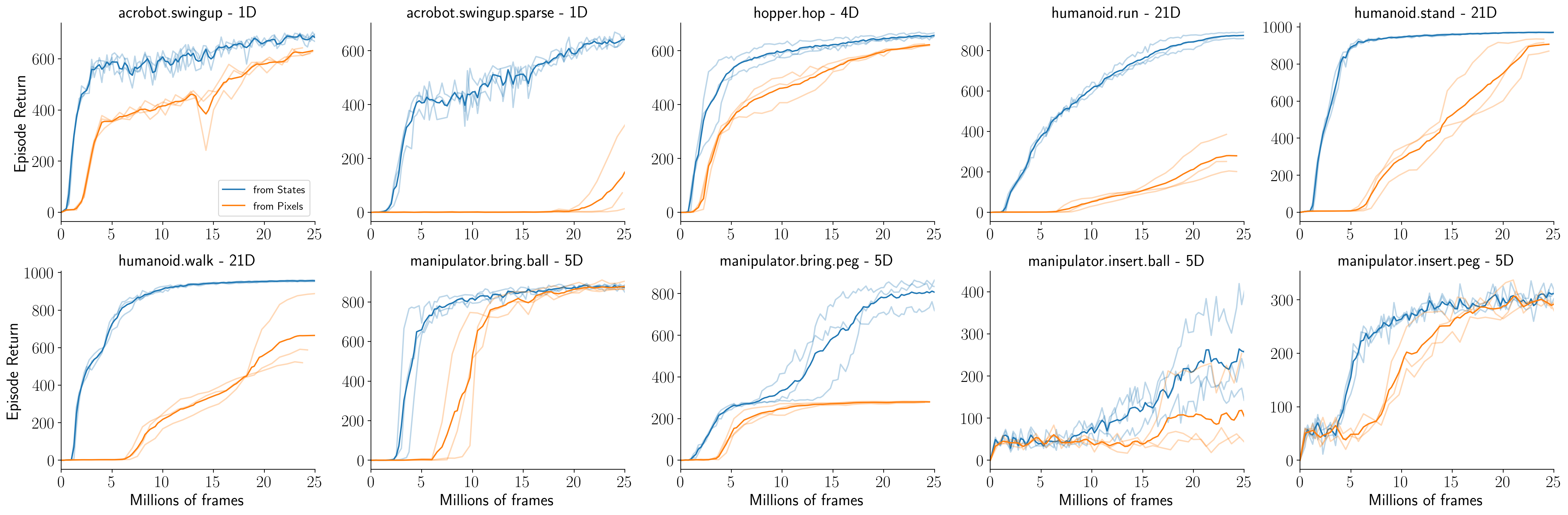}
\vspace*{-7mm}
\caption[]{
\label{fig:hard-results-seeds}
\textbf{Reproducibility of \smuzero{} from state and raw pixel inputs on the hard and manipulator tasks.} Performance of \smuzero{} (3 seeds per experiment) throughout training on the DM Control Humanoid Run task.
}
\end{figure*}

In order to evaluate the reproducibility of \smuzero{} from state inputs and raw pixel inputs, we show the individual performance of 3 seeds on the hard and manipulator tasks in Figure \ref{fig:hard-results-seeds}. Overall, the variation in performance across seeds is minimal.

In addition, we show the individual performance of 6 seeds when sampling $K=3,5,10,20,40$ actions on the \emph{humanoid.run} task. We observe that even when the number of samples is small, performance stays very reproducible across runs.

\begin{figure*}
\includegraphics[width=\textwidth]{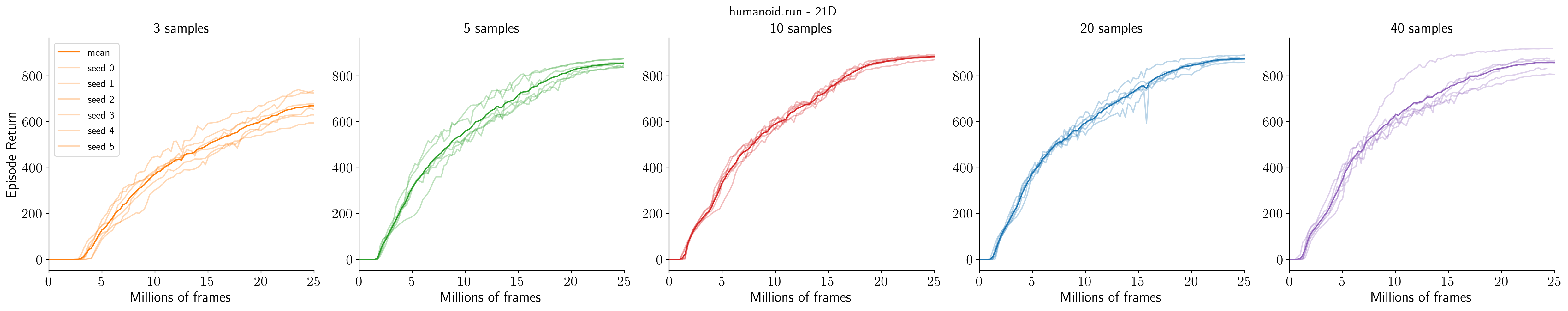}
\vspace*{-7mm}
\caption[]{
\label{fig:control-suite-reproducibility}
\textbf{Reproducibility of \smuzero{} on the \emph{humanoid.run} task with 3, 5, 10, 20 and 40 action samples.} Performance of \smuzero{} (6 seeds per experiment) throughout training on the DM Control Humanoid Run task.
}
\end{figure*}

\subsection{Ablation on using $\hat{\pi}_\sample$ vs $\pi$}
We evaluated the practical importance of using $\hat{\pi}_\sample=\hat{\sample}/\sample \pi$ instead of just $\pi$ in \smuzero{}'s PUCT formula and ran experiments on the \emph{humanoid.run} task.

We expect that as the number of samples increases, the difference will go away as $\lim_{K\to\infty} \hat{\pi}_\sample = \lim_{K\to\infty} \hat{\sample}/\sample \pi = \pi$. We therefore looked at the difference in performance when drawing $K=5$ or $K=20$ samples.

Furthermore, evaluating the Q values of all sampled actions at the root of the search tree before the start of the search puts more emphasis on the values and less on the prior in the PUCT formula. We therefore also show the difference in performance with and without Q evaluations (\emph{no Q} in the figure).

\begin{figure}
\includegraphics[width=\columnwidth]{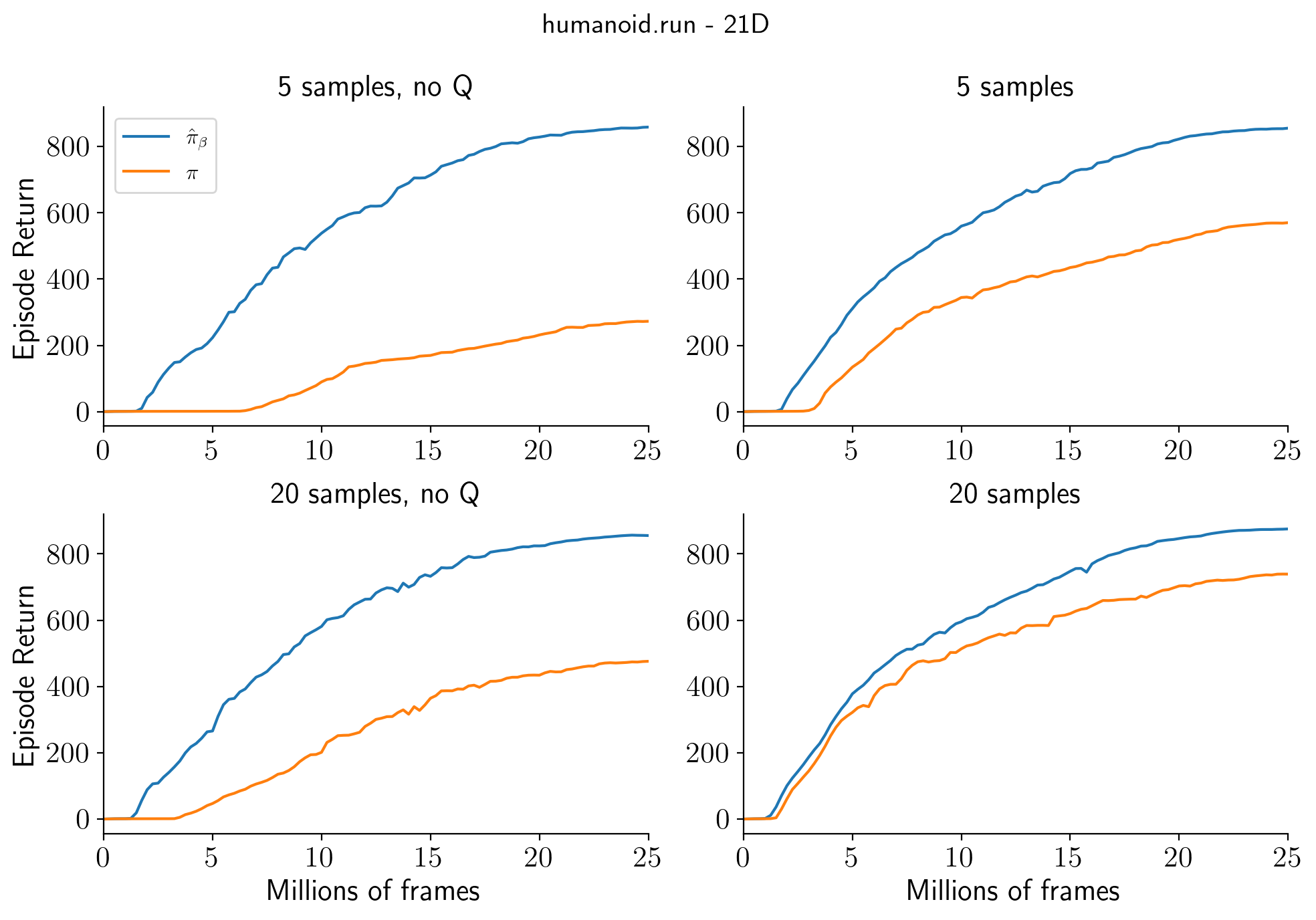}
\vspace*{-7mm}
\caption[]{
\label{fig:control-suite-ptilde}
\textbf{Performance of \smuzero{} using $\hat{\pi}_\sample$ vs $\pi$ on the \emph{humanoid.run} task.} Performance of \smuzero{} (3 seeds per experiment) throughout training on the DM Control Humanoid Run task evaluated with $K=5$ or $K=20$ samples and with or without (no Q) evaluating the Q values of all sampled actions at the root of the search tree. It is much more robust to use $\hat{\pi}_\sample$ over $\pi$ in \smuzero{}.
}
\end{figure}

The experiments in Figure \ref{fig:control-suite-ptilde} confirm that it is much better to use $\hat{\pi}_\sample$ when the number of samples is small and not  evaluating the Q values. The performance drop of using $\pi$ is attenuated by evaluating the Q values at the root of the search tree, but it is still better to use $\hat{\pi}_\sample$ even in that case.

\section{Go Experiments}
For the Go experiments, we mostly used the same neural network architecture, optimisation and hyperparameters used by \muzero{} \cite{muzero} with the following differences. Instead of using the outcome of the game to train the value network, we used n-step bootstrapping with $n=25$ where the value used to bootstrap was the averaged predictions of a target network applied to $4$ consecutive states at indices $n+i$ for $i\in\left[0, 3\right]$. We averaged multiple consecutive target network value predictions due to the alternation of perspective for value prediction in two-player games; using the average of multiple estimates ensures that learning is based on the estimates for both sides. We observed that this reduced value overfitting and allowed us to train \muzero{} while generating less data. In addition, we used a search budget of $400$ simulations per move instead of $800$ in order to use less computation.

We evaluated the network checkpoints of \muzero{} and \smuzero{} throughout training playing $100$ matches with a search budget of $800$ simulations per move. We anchored the Elo scale to a final \muzero{} baseline performance of $2000$ Elo.

\section{Atari Experiments}
For the Atari experiments, we used the same architecture, optimisation and hyperparameters used by \muzero{} \cite{muzero}.

We evaluated the network checkpoints of \muzero{} and \smuzero{} throughout training playing $100$ games with a search budget of $50$ simulations per move.

\section{Search}
\label{search}

The full PUCT formula used in \smuzero{} is:
$$arg\,max_a Q(s,a)+c(s) \cdot (\frac{\hat{\sample}}{\sample}\pi)(s, a)\frac{\sqrt{\sum_b N(s,b)}}{1+N(s,a)}$$
where
$$
c(s) = c_1 + \log\frac{1+c_2+\sum_b N(s,b)}{c_2}
$$

with $c_1 = 1.25$ and $c_2 = 19652$ in the experiments for this paper. Note that at visit counts $N(s)=\sum_b N(s,b) \ll c_2$, the $log$ in the exploration term is approximately $0$ and the formula can be written:
$$arg\,max_a Q(s,a) + c_1 \cdot (\frac{\hat{\sample}}{\sample}\pi)(s, a)\frac{\sqrt{\sum_b N(s,b)}}{1+N(s,a)}$$

\section{Sample-based Policy Improvement and Evaluation Proofs}
\label{proofs}
\textbf{Lemma}. $\hat{Z}_\sample$ and $Z$ are linked by:
$$\lim_{K\to\infty} \hat{Z}_\sample = Z$$

\textbf{Proof}
$\hat{Z}_\sample(s)$ is defined such that $\sum_{a\in\mathcal{A}}(\hat{\sample}/\sample)(a|s) f(s, a, \hat{Z}_\sample(s))=1$.

Therefore
\begin{equation*}
\begin{split}
1 =& \lim_{K\to\infty} \sum_{a\in\mathcal{A}} (\hat{\sample}/\sample)(a|s)  f(s, a, \hat{Z}_\sample(s)) \\
=& \lim_{K\to\infty} \sum_{a\in\mathcal{A}} f(s, a, \hat{Z}_\sample(s))
\end{split}
\end{equation*}
where we used $\lim_{K\to\infty} \hat{\sample} = \sample$ to go from line 1 to 2.

We therefore have
$$\lim_{K\to\infty} \sum_{a\in\mathcal{A}} f(s, a, \hat{Z}_\sample(s))=1=\sum_{a\in\mathcal{A}} f(s, a, Z(s))$$
which shows by the uniqueness of $Z$ that $\lim_{K\to\infty} \hat{Z}_\sample = Z$.

\textbf{Theorem}. For a given random variable $X$, we have
$$\mathbb{E}_{a\sim\ipi}[X|s] = \lim_{K\to\infty} \sum_{a\in\mathcal{A}} \ibpi(a|s) X(s, a)$$

Furthermore, $\sum_{a\in\mathcal{A}} \ibpi(a|s) X(s, a)$ is approximately normally distributed around $\mathbb{E}_{a\sim\ipi}[X|s]$ as $K\to\infty$:
$$\sum_{a\in\mathcal{A}} \ibpi(a|s) X(s, a) \sim \mathcal{N}(\mathbb{E}_{a\sim\ipi}[X|s], \frac{\sigma^2}{K})$$
where $\sigma^2 = Var_{a\sim\beta}[\frac{f(s, a, Z(s))}{\sample} X(s, a)|s]$.

\textbf{Proof}. We have
\begin{equation*}
\begin{split}
&  \mathbb{E}_{a\sim\ipi}[X(s,a)|s] \\
&= \mathbb{E}_{a\sim\sample}[(\ipi/\sample)(a|s) X(s,a)|s] \\
&= \mathbb{E}_{a\sim\sample}[f(s, a, Z(s))/\sample(a|s) X(s,a)|s] \\
&= \lim_{K\to\infty} \sum_{a\in\mathcal{A}} (\hat{\sample}/\sample)(a|s) f(s, a, Z(s)) X(s,a) \\
&= \lim_{K\to\infty} \sum_{a\in\mathcal{A}} (\hat{\sample}/\sample)(a|s) f(s, a, \hat{Z}_\sample(s)) X(s,a)\\
&= \lim_{K\to\infty} \sum_{a\in\mathcal{A}} \ibpi(a|s) X(s,a)
\end{split}
\end{equation*}
where we used the law of large numbers to go from line 2 to 3, replacing the expectation with the limit of a sum, and the lemma to go from line 3 to 4.

Using the central limit theorem from line 2, we can also show that as $K\to\infty$,
$$\sum_{a\in\mathcal{A}} (\hat{\sample}/\sample)(a|s) f(s, a, Z(s))X(s, a) \to \mathcal{N}(\mathbb{E}_{a\sim\ipi}[X|s], \frac{\sigma^2}{K})$$
in distribution with $\sigma^2 = Var_{a\sim\beta}[\frac{f(s, a, Z(s))}{\sample} X(s, a)|s]$.

Making the approximation of swapping in $\hat{Z}_\sample$ for $Z$ based on the lemma, we obtain that as $K\to\infty$:
$$\sum_{a\in\mathcal{A}} \ibpi(a|s) X(s, a) \sim \mathcal{N}(\mathbb{E}_{a\sim\ipi}[X|s], \frac{\sigma^2}{K})$$

\textbf{Corollary}.
The sample-based policy improvement operator converges in distribution to the true policy improvement operator:
$$\lim_{K\to\infty} \ibpi = \ipi$$

Furthermore, the sample-based policy improvement operator is approximately normally distributed around the true policy improvement operator as $K\to\infty$:
$$\ibpi(a|s) \sim \mathcal{N}(\ipi(a|s), \frac{\sigma^2}{K})$$
where $\sigma^2 = Var_{a\sim\beta}[\frac{f(s, a, Z(s))}{\sample} \mathbbm{1}(a)|s]$.

\textbf{Proof}. We obtain the corollary by using $X(s,a)=\mathbbm{1}(a)$ in conjunction with
$\ipi(a|s)= \mathbb{E}_{a\sim\ipi}[\mathbbm{1}(a)|s]$ and $\ibpi(a|s)=\sum_{b\in\mathcal{A}} \ibpi(s, b) \mathbbm{1}(a)$

\section{The \muzero{} Policy Improvement Operator}
\label{mz-pi}
Recent work \cite{grill2020montecarlo} showed that \muzero{}'s visit count distribution was tracking the solution $\bar{\pi}$ of a regularised policy optimisation problem:
$$\bar{\pi} = arg\,max_\Pi  Q^T\Pi - \lambda_N KL(\pi, \Pi)$$
where KL is the Kullback–Leibler divergence and $\lambda_N$ is a constant dependent on $c$ and the total number $N$ of simulations.

$\bar{\pi}$ can be computed analytically:
$$\bar{\pi}(a|s) = \lambda_N \frac{\pi(s,a)}{Z(s) - Q(s,a)}$$
where $Z(s)$ is a normalising factor such that $\forall a\in\mathcal{A}, \bar{\pi}(a|s)\geq 0$ and $\sum_{a\in\mathcal{A}} \bar{\pi}(a|s)=1$.

In other words, using the terminology introduced in Section \ref{sample-based-policy-improvement}, \muzero{}'s policy improvement can be approximately written:
$$\ipi(a|s) \approx f(s, a, Z(s))$$
where
$$f(s, a, Z(s)) = \lambda_N \frac{\pi(a|s)}{Z(s) - Q(s,a)}$$
and is therefore action-independent.

Let's consider the visit count distribution $\ihpi$ obtained by searching using prior $\hat{\pi}_\sample=\hat{\sample}/\sample \pi$.

Using \cite{grill2020montecarlo}, we can write:
\begin{equation*}
\begin{split}
\ihpi(s, a)
& \approx \lambda_N \frac{\hat{\pi}_\sample(a|s)}{\hat{Z}_\sample(s) - Q(s,a)} \\
& = \lambda_N \frac{(\hat{\sample}/\sample \pi)(a|s)}{\hat{Z}_\sample(s) - Q(s,a)} \\
& = (\hat{\sample}/\sample)(a|s) f(s, a, \hat{Z}_\sample(s))
\end{split}
\end{equation*}
where $\hat{Z}_\sample(s)$ is such that $\forall a\in\mathcal{A}, \ihpi(s, a)\geq 0$ and $\sum_{a\in\mathcal{A}} \ihpi(s,a)=1$.

This shows that $\ihpi$ is the action-independent sample-based policy improvement operator associated to $\ipi$.

\end{document}